\documentclass[preprint]{vgtc}               

\graphicspath{{figures/}{pictures/}{images/}{./}} 
\usepackage{balance}

\usepackage{times}                     

\usepackage{tabu}                      
\usepackage{booktabs}                  
\usepackage{lipsum}                    
\usepackage{mwe}                       
\usepackage{makecell}
\usepackage{mathptmx}   
\usepackage{amsmath,amssymb,amsfonts}
\usepackage{bm}
\usepackage{soul}
\usepackage{algorithmic}
\usepackage{graphicx}
\usepackage[thicklines]{cancel}
\onlineid{1338}

\vgtccategory{Research}

\vgtcinsertpkg

\preprinttext{This is the author’s version of the article. To appear in an IEEE ISMAR conference.}

\title{MIRAGE: Multimodal Intention Recognition and Admittance-Guided Enhancement in VR-based Multi-object Teleoperation}

\author{Chi Sun\thanks{e-mail: celeste-sora.sun@connect.polyu.hk}\\
        \scriptsize The Hong Kong Polytechnic University
\and Xian Wang\thanks{e-mail: xiann.wang@connect.polyu.hk}\\
        \scriptsize The Hong Kong Polytechnic University
\and Abhishek Kumar\thanks{e-mail: abhishek.k.kumar@jyu.fi}\\
        \scriptsize University of Jyv\"{a}skyl\"{a}
\and Chengbin Cui\thanks{e-mail: chengbin.cui@connect.polyu.hk}\\
        \scriptsize The Hong Kong Polytechnic University
\and Lik-Hang Lee\thanks{e-mail: lik-hang.lee@polyu.edu.hk (the corresponding author)}\\
        \scriptsize The Hong Kong Polytechnic University}

\teaser{
  \centering
  \includegraphics[width= 0.85\linewidth]{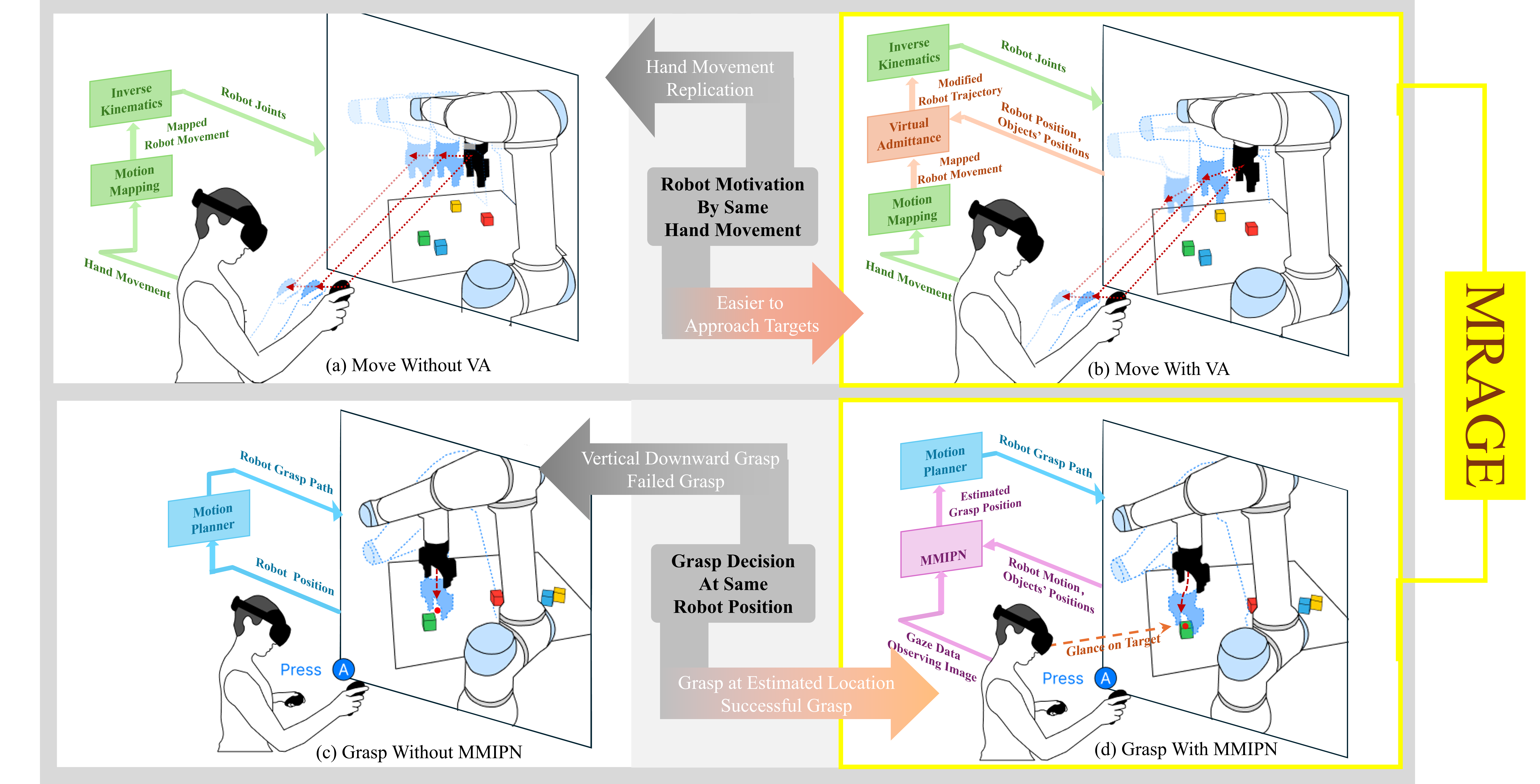}
  \caption{A pictorial description of the \textit{MIRAGE} framework that enhances HRI tele-grasping capability for multiple objects in VR. \textcolor{black}{\textit{MIRAGE}} divides the multi-object grasping task into two phases: movement (manual) and grasping (semi-automatic). Each phase has a specific assistance method designed in \textit{MIRAGE}: In the movement (manual) phase, Virtual Admittance (VA) modifies the robot trajectory (b), comparing to the non-VA condition (a), is easier to motivate the robot to approach target  \textcolor{black}{through the} same hand movement; in the grasping (semi-automatic) phase, a Multimodal-CNN-based Human Intention Perception Network (MMIPN) is proposed to estimate the human desired grasp position for robot grasp motion plan (d), and the non-MMIPN condition plans the grasping motion as a vertical downward path (c).}
  \label{fig:teaser}
}

\abstract{
Effective human-robot interaction (HRI) in multi-object teleoperation tasks faces significant challenges due to perceptual ambiguities in virtual reality (VR) environments and the limitations of single-modality intention recognition. This paper proposes a shared control framework that combines a virtual admittance (VA) model with a Multimodal-CNN-based Human Intention Perception Network (MMIPN) to enhance teleoperation performance and user experience. The VA model employs artificial potential fields to guide operators toward target objects by adjusting admittance force and optimizing motion trajectories. MMIPN processes multimodal inputs—gaze movement, robot motions, and environmental context—to estimate human grasping intentions, helping overcome depth perception challenges in VR. Our user study evaluated four conditions across two factors, and the results showed that MMIPN significantly improved grasp success rates, while the VA model enhanced movement efficiency by reducing path lengths. Gaze data emerged as the most crucial input modality. These findings demonstrate the effectiveness of combining multimodal cues with implicit guidance in VR-based teleoperation, providing a robust solution for multi-object grasping tasks and enabling more natural interactions across various applications in the future.
} 

\keywords{Human-robot interaction, multimodal CNN, virtual reality, shared control}

\begin{document}


\firstsection{Introduction}
\maketitle
With the rapid development of robotics and metaverse technology, in particular, teleoperation technology has brought diverse modes and expanded opportunities for remote operations. In the fields of aerospace manipulator operation~\cite{goza2004CHItelepresence, ma2023TAEShuman}, extraterrestrial ground exploration~\cite{arzo2022TAESessential}, nuclear environment maintenance~\cite{marturi2016towards,chiou2022IROSnuclear}, remote medical surgery~\cite{wagner2007TROsurgery, Burgner2014TMECHsurgery}, and life care assistance~\cite{lyu2020TEHMhomecare}, teleoperation already has a wide range of technical needs and successful application experience. 
The rise and prosperity of Metaverse technology have promoted the applications of virtual reality (VR) in industrial teleoperation~\cite{whitney2019comparing,barentine2021HRIvr,nenna_enhanced_2023}. 
The immersion of VR can provide a more realistic experience for the teleoperation.

VR-based human-robot interaction (HRI) is more natural and efficient than traditional input methods (e.g., buttons and keyboards) through intuitive spatial mapping~\cite{groechel2022RAMvatoolhri,nenna_enhanced_2023}, where controller movements in virtual space directly correspond to robot actions in physical space, creating a more natural and cognitively efficient control paradigm~\cite{lin2020ICRAshared, Darvish2023TROteleoperationsurvey}. This method enables complete operational autonomy to operators but presents significant challenges, including the absence of haptic feedback and reduced accuracy due to gross human movements, potentially affecting the efficiency of decisions.
In addition to this, the ambiguity and uncertainty of human behavior 
in HRI increase the difficulty of executing high-precision teleoperation tasks~\cite{yoojin2021IROSarbitratehri,li2024RALconflicthri}. Therefore, the shared control framework, ``the case where both automation and humans work on the same task and at the same time''~\cite{sheridan1978human}, enables the robot to autonomously solve basic control safety and stability issues when humans convey control instructions related to commands to the robot~\cite{li2023ThapticSharedcontrolsurvey}.

Intention recognition and seamless interaction are key factors for the effect of shared control~\cite{fonooni2015applyingsharedcontrol, aronson2024HRIintentionalsharedcontrol}. In existing shared control, human intention is usually interpreted as parameters in the model after physical modeling, and seamless interaction methods are most often implemented by assigning weights or separating channels of the shared control framework~\cite{luo2024user, krupke2016immersive,xu2022shared}, but such methods are often adapted to physical HRI with haptic feedback and single-target interaction tasks. In teleoperation schemes implemented using VR environments, it is usually expected that after reproducing a realistic remote environment, the slave robot can complete complex tasks remotely, such as accurately reaching and operating target objects in a multi-object environment. A typical task for this manipulation requirement is multi-object tele-grasping, which plays a significant role in complex tasks across various domains~\cite{marturi2016towards}. This task requires the HRI system can both compensate for the lack of human spatial perception and accurately interpret human intent.

Therefore, in our study, we focused on three research questions in utilizing a shared control framework in VR-based HRI for the multi-object tele-grasping task: \textbf{RQ1}: How to provide intuitive guidance for the user's spatial movement under the direct shared control framework without the force feedback?  \textbf{RQ2}: How to realize the precise human intention recognition among multiple objects? \textbf{RQ3}: How to integrate the human intention and robot autonomy in the HRI scheme to enhance grasping task performance among multiple objects? 

To address these questions, we present \textit{MIRAGE}: a multimodal intention recognition and virtual-admittance-guided enhancement framework for  VR-based multi-object tele-grasping. 
We implemented the \textcolor{black}{\textit{MIRAGE}} design on an experimental platform and conducted a user study involving 16 participants. The key contributions in our work are: (1) For \textbf{RQ1}, we designed an implicit Virtual Admittance (VA) model into a VR-based shared control framework based on VR without haptic feedback; this strategy dynamically adjusts the virtual admittance forces in real time during the HRI process. This reshapes the robot's trajectory in VR as a kind of visual cue, guiding operators toward target objects. 
(2) For \textbf{RQ2}, we proposed a novel human \textcolor{black}{intention recognition} method (MMIPN) through Convolutional Neural Networks (CNN). The network utilizes gaze, robot motions, and environmental context (observing image and potential target positions)  to estimate human intended grasping position. Since the combination of modalities in MMIPN is underexplored, a multimodal grasping intent dataset is created from data collected by the experimental platform. We implemented an ablation study using this dataset to analyze the impact of each modality, found the gaze modality to have the most significant effect and verified that the complete MMIPN had the highest estimation accuracy.
(3) For \textbf{RQ3}, the \textit{MIRAGE} framework is suggested in this paper, integrating VA and MMIPN in different grasping phases, movement and grasping to balance the human intention and robot autonomy towards improving task performance. We conducted the user study across four different conditions: a baseline (no assistance), VA only, MMIPN only, and a combination of both. Key results demonstrate that the MMIPN method significantly improved task completion performance. Additionally, the VA's assistance notably increased the operator's movement efficiency.

\section{Related Works}
\paragraph{Teleoperations in Virtual Reality Interface}

Teleoperation relies on visual feedback to provide the operator with clues to guide the robot's operation, and an immersive virtual reality interface is considered an effective remote environment display for teleoperation~\cite{rodehutskors2015intuitive,stoyanov2018IROSassisted}. Earlier VR displays were designed to create a 3D remote scene through software rendering, enabling remote operators to gain a three-dimensional spatial perspective. For instance, David et al.,~\cite{kent2017comparison} developed an AR-based virtual reconstruction environment that offered a 3D drag-and-click interactive interface for remote grasping tasks on a 2D screen. However, users still need to access 3D information on a 2D screen, which lacks immersion, and the interaction is not intuitive.

In contrast, head-mounted displays (HMDs) liberate users from these constraints, allowing for a more realistic understanding of the remote environment \cite{walker2023virtual,chang2022virtual}. With the support of somatosensory controllers in VR, user intuitiveness and operational freedom are significantly enhanced. VR feedback in immersive HMD has been proven to be effective in improving the efficiency of human-robot teleoperation~\cite{whitney2019comparing, barentine2021HRIvr, wonsick2020systematic}. For example, Wan et al.~\cite{wan2024virtual} employed gloves and handles to control remote industrial robotic arms in VR, creating a more immersive and personalized platform for battery removal operators.
The existing HRI interface in virtual reality has studied a variety of interactive interfaces, including the traditional mouse and keyboard~\cite{lemasurier2024ROMANcomparing}, virtual button~\cite{sun2020new}, the fixed controller~\cite{franzluebbers2019remote,abi2019haptic}, the wireless tracking controller~\cite{xu2022shared,jin2024ISMARmitigating}, the wearable glove~\cite{wan2024virtual}, body (hand) tracking~\cite{krupke2016immersive, lin2020ICRAshared, sun2020new}, and gesture~\cite{wang2024robotic}. Motion mapping is a more intuitive method of teleoperation~\cite{lin2020ICRAshared, nenna_enhanced_2023}, but the lack of force feedback is a significant concern in this kind of method. 
Haptic feedback can effectively bridge the gap in distance perception through guidance~\cite{sun2020new}, while studies have shown that wireless controller vibration does not affect error cancellation~\cite{jin2024ISMARmitigating}. Therefore, \textcolor{black}{\textit{MIRAGE}} proposed to apply a VA model as an implicit assistance in virtual reality to compensate for the lack of distance and position-aware feedback. This method could reshape the robot's trajectory by setting up virtual-force-based force-position hybrid control to produce intuitive visual cues.

\paragraph{Assistance in HRI Tele-Grasping}
Robot teleoperated grasping plays a significant role in accomplishing complex tasks across various fields, especially in hazardous scenarios~\cite{marturi2016towards}. Existing research focuses on how to provide effective HRI assistance in the context of single-object grasping. For instance, Adjigble et al. provide guidance on the grasping posture for manipulating a single object by predicting grasping postures using local contact torque within a shared control framework, accompanied by tactile feedback through a 6-DOF force feedback device~\cite{adjigble2019IROSassisted, adjigble2021IORSspectgrasp}. Eltan Bounly et al. studied assisting in approaching the target through tactile force feedback gloves and allocating position-based shared control weights~\cite{eltanbouly2022assistive}. Stoyanov et al. divide single-object grasping into two sub-phases, alignment and grasping, and categorize the control tasks into three distinct levels of autonomy; however, task switching is implemented in the acceleration space through decision variables~\cite{stoyanov2018IROSassisted}. These works are all achieved under the condition that the target manipulation object is single and its position is clear, and rely to a certain extent on haptic feedback.  In contrast, applying these single-object manipulation methods when confronted with multi-object scenarios lacks an important prerequisite, which is the clarity of the target object. 

Therefore, in multi-object grasping, alignment becomes particularly crucial for clearly identifying the manipulated target~\cite {stoyanov2018IROSassisted}. It is worth mentioning that 3D UIs in VR have transitioned from 2D interfaces, and users have to operate the aforementioned steps in 3D spaces. As such, a significant challenge in teleoperated grasping remains the difficulty in estimating depth information in VR~\cite{arevalo2021CHIassisting,zhu2023shared}. 
 In other words, the depth error of the operator's visual observation will cause ambiguity in estimating the 3D object being manipulated. As such, the target inference based on the manipulated position is usually unreliable~\cite{abi2019haptic}. Therefore, in multi-target grasping without haptic feedback, we propose \textcolor{black}{\textit{MIRAGE}} to enhance the perception and approach of the operator to the multi-target position by solely visual cues (i.e., the virtual admittance method) and then further inference of the grasping target through intention recognition by leveraging multiple perceptions in the VR interface, including users' gaze information.

\paragraph{Human Intention Recognition in HRI}
In the HRI teleoperation, human intention represents a variety of information, including the trajectory or direction of the robot movement that humans expect~\cite{sun2024development, song2024robot, Nara2023ICRAintentpredict}, the object or manipulation position that humans expect the robot to operate~\cite{YANG2023Gaze,li20173}, the operation that humans expect the robot to perform on the object~\cite{de2015RASrecognizing, Li2014Gesture,dutta2016predicting}, or the predefined action that humans expect the robot to complete~\cite{abuduweili2019adaptable}. In the multi-target grasping task we discussed, we define human intention as the object and its position that humans intend to grasp.
Gaze cues are effective in expressing human intentions in multi-object search tasks in 2D~\cite{pan2024gaze} and 3D scenarios~\cite{YANG2023Gaze,li20173}. EEG~\cite{lin2023neuralEEG}
 and visual context~\cite{yang2023figgaze, molina2024asymmetric} are introduced into the gaze-based framework to enhance the robustness of human intention recognition in multi-object grasping, and deep neural networks are used to fuse multimodal features to obtain accurate human intention. However, these methods are essentially entirely reliant on gaze-driven approaches, primarily conducting intention inference research focused on individuals with disabilities or the elderly with mobility impairments, in which cases, relying on a single modality often falls short of accuracy and can increase the cognitive load on operators \cite{li20173, plopski2022CSUReye}. In addition, the motion of the human body is absent in HRI in these cases. 
 In the human-robot shared control teleoperated scheme, the guidance movements from humans to robots also serve as important cues of human intentions~\cite{schydlo2018ICRAmultimodalgaze}. We incorporate the end-effector position of the robot under direct shared control into the modality range required for intention inference and propose a human intention recognition method aimed at multi-object manipulation that integrates modalities, including gaze, environmental context, and robot positions.

\section{MIRAGE: Framework Overview}
This framework focuses on a task that involves placing a robot arm with a gripper in a remote environment containing several manipulable objects. The operator uses a VR headset and controller as perception and control devices to drive the remote robot to pick up the correct object according to the operator's intention.

\subsection{VR Teleoperation Framework}\label{shaed_control}
The robot's base coordinate frame $\{\bm{R}\}$ is regarded as the right-handed robot's reference coordinate frame on the remote side, and $\bm{x}_r \in \mathbb{R}^{3}$ is the robot's end position. On the local side, the operator's controller grasping position is defined in the left-handed world coordinate frame $\{\bm{H}\}$ as $\bm{x}_h \in \mathbb{R}^{3}$, and the end position of the virtual robot in the local VR environment is defined in the virtual robot' base coordinate frame $\{\bm{V}\}$ as $\bm{x}_v \in \mathbb{R}^{3}$. The y- and z-axes of the two world coordinate frames are aligned.

Robot systems are commonly controlled in Cartesian space for manipulation tasks, where position control is employed and robot joint angles are computed via inverse kinematics. We applied an incremental direct teleoperation control method to enlarge the operator's workspace by switching the control status through his controller. With the control switch function $\bm{\aleph} = \{0 \quad (\rm{disable\quad telecontrol)} \, | \, 1 \quad \rm{(enable \quad telecontrol)}\}$ ~\cite{FU2025TROcontactrichsharedcontrol} , the remote robot position can be expressed as 
\begin{equation}\label{equ:telemapping}
\bm{x}_r(t) = \bm{x}_r(t_0) + \bm{\aleph} \int_{t_0}^{t_s} k_m T^r_h \dot{\bm{x}}_h dt,
\end{equation}
where $k_m$ is the scaling parameter, $t_0$ and $t_s$ are the time instant of the teleoperation switch on and off; $\dot{\bm{x}}_h \in \mathbb{R}^{3}$ is the operator's hand movement velocity, $\bm{x}_r(t_0) \in \mathbb{R}^{3}$ is the robot's initial positions and $\bm{x}_r(t) \in \mathbb{R}^{3}$ is its desired position after the task execution. $T^r_h \in \mathbb{R}^{3 \times 3}$ is the transformation matrix from $\{\bm{H}\}$  to {$\{\bm{V}\}$}.

The rotation $\bm{\omega}_v$ in the virtual engine uses the left-handed ZYX rotation order to represent the Euler angle, while the real robot operating system normally uses the right-handed ZYX rotation $\bm{\omega}_r$ order to represent the Euler angle. Therefore, based on the uniqueness of the rotation quaternion, the quaternion is used as a bridge to unify the conversion of the coordinate frame and then solve the Euler angle through the rotation order.
The relationship between the rotation quaternion $q_v = \{w_v,a_v,b_v,c_v\}^T$ in $\{\bm{V}\}$ and the rotation quaternion $q_r = \{w_r,a_r,b_r,z_r\}^T$ in $\{\bm{R}\}$ is described as  \textcolor{black}{$q_v = T^r_v q_r$, where $T^r_v = \rm{diag}(1,1,-1,-1)$ is the diagonal transformation matrix.}

\subsection{Robot Kinematics}\label{IK}
Human operators are more cautious when grasping an object, which limits the robot's motion to three degrees of freedom. Our system and methodology have no restrictions on tandem robotic arms with 6 DOF or tandem redundant robotic arms with greater than 6 DOF. \textcolor{black}{While parallel or hybrid robot arms are not included in the proposed framework, this is due to their kinematics calculations differing from those of serial robot arms.}
 The angles of the n-DOF ($n \geq 6$) robot $\bm{\Theta }= (\theta_1, \theta_2,\dots,\theta_n)$ were calculated by the robot kinematics through the desired position and rotation of the robot end as
$ \mathbf{T}_{\text{end}}(\bm{x}_r,\bm{\omega}_r) = \notag
\mathbf{T}_{0} \cdot \mathbf{T}_{1}(\theta_1) \cdot \mathbf{T}_{2}(\theta_2) \cdots \mathbf{T}_{n}(\theta_n)$
, where the $\mathbf{T}_i(\theta_i),i\in[1,n]$ is the Homogeneous rotation transformation matrix based on the joint angle $\theta_i $ and the $\mathbf{T}_{\text{end}}(\bm{x}_r,\bm{\omega}_r) \textcolor{black}{\in\mathbb{R}^{4\times4}}$ is the Homogeneous transformation matrix consisted of the rotation matrix $R_r \textcolor{black}{\in\mathbb{R}^{3\times3}}$ and the translation vector $P_r \textcolor{black}{\in\mathbb{R}^{3}}$. 

The solution set of the inverse kinematics angles is obtained by the \textcolor{black}{ analytical solution \cite{Kannade2010PHDTHESISiksolution} through the robot's inverse kinematics model}, and the desired solution $\bm{\Theta}$ is selected according to the shortest distance principle. 
In our teleportation framework, the joint angle commands are given via this control  \textcolor{black}{signal} $\mathbf{T}_{\text{end}}(\bm{x}_r,\bm{\omega}_r)$ \textcolor{black}{, which} simultaneously drive the motions of both the physical robot in the remote setting and its virtual counterpart in virtual reality, ensuring synchronized behavior across both realities. 

    \subsection{Virtual Admittance Model}\label{admittancecontrol}
Direct teleoperation is a leader-follower tracking process of expectations generated by the remote human control input. However, during the grasping process, the human operator's operating intention is usually obvious, that is, approaching an object to further perform the operation. Therefore, we design a virtual admittance structure to combine human control intention and robot autonomy:
\begin{equation}
\label{impedance}
M_d^r \ddot{\bm{e}}_d^r + C_d^r \dot{\bm{e}}_d^r +K_d^r\bm{e}_d^r = \bm{F}^r_d
\end{equation}
where $\bm{e}^r_d=\bm{x}_r - \bm{x}_d$, $\bm{x}_d$ is the desired trajectory which is motivated by the human motion from the leader side and the $\bm{x}_r$ is the mixed trajectory for the end of the robot arm. $M_d^r \in\mathbb{R}^{3\times3}$ is the mass parameter matrix, $C^r_d \in\mathbb{R}^{3\times3}$ is the damping parameter matrix, $K^r_d \in\mathbb{R}^{3\times3}$ is the spring parameter matrix.

The Artificial Potential Field (APF) is established at each point of the object in the remote scene to use the position information of the potential target of interest.  At each object $o_i$ with a position of $P_{oi}$, its APF model is

\begin{equation}\label{equ:apfmodel1}
\bm{F_i} = k_i\cdot\frac{P_r-P_{oi}}{\|P_r-P_{oi} \|^2},
\end{equation}
where the $k_i$ is the strength factor of the attractive force. Considering $n$ objects in the scene, the virtual control force in the  ~\cref{impedance} is  $\bm{F}^r_d = \sum\limits_{i=1}^{n}\bm{F_i}$.

\subsection{Multimodal-CNN-based Human Intention Recognition}\label{MMIPN}
To accurately implement human grasping expectations, the paper proposes a CNN-based \ul{M}ulti\ul{M}odal human \ul{I}ntention \ul{P}erception \ul{N}etwork (MMIPN). MMIPN leverages multimodal human information, i.e., the human gaze and visual observation, and the scene content, including the robot motion and the potential objects, to estimate the position of the target object that humans expect.

\begin{figure}[h]
\centering
\includegraphics[width=.85\columnwidth]{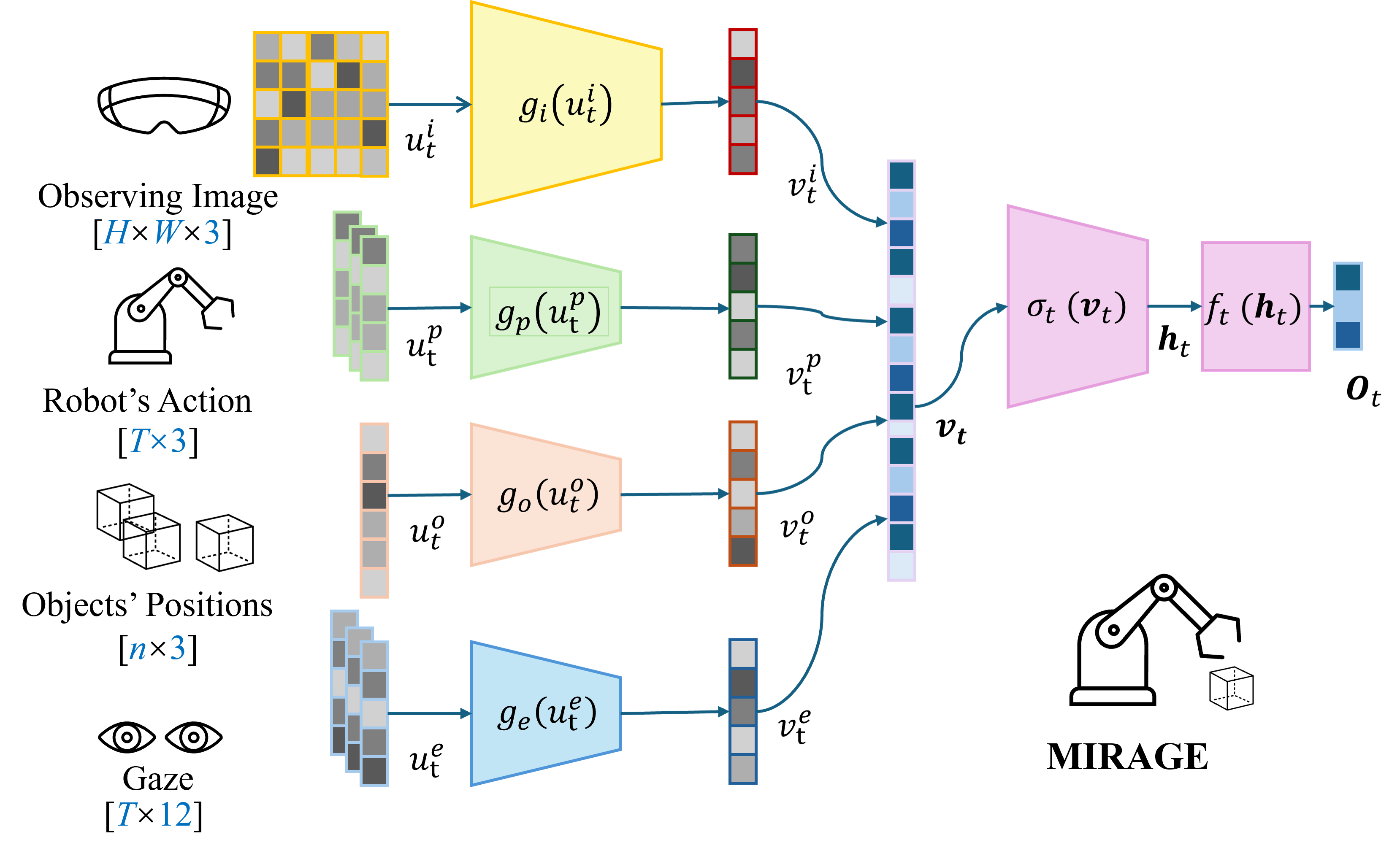}
\caption{The brief MMIPN structure: The network received four channel inputs, which are (1) the observing image \textcolor{black}{$u_t^i$}, (2) the robot's action \textcolor{black}{$u_{\rm{t}}^p$}, (3) the potential objects' positions \textcolor{black}{$u_t^o$}, and (4) the gaze data \textcolor{black}{$u_{\rm{t}}^e$}. After the multimodal data undergoes feature extraction by the CNN network, it is spliced into a complete compression vector. Then, through the fully connected layer, it is activated by the activation function to estimate the target position as the human intent.} 
\label{fig_cnn}
\end{figure}

\cref{fig_cnn} illustrates the structure of MMIPN. The input vector of the network is designed with four modalities as 

  \begin{subequations}
\begin{align}
         &\bm{I_t} = (u^i_t, u^p_\mathbf{t}, u^o_t,u^e_\mathbf{t}) \\
     &\{ \mathbf{t} | t-T +1 \leq \mathbf{t} \leq t \},
\end{align}
 \end{subequations}
 \textcolor{black}{$u^i_t \in \mathbb{R}^{H \times W \times 3}$ and $u^o_t \in \mathbb{R}^{n \times 3}$ are the vectors representing the input of the image feature and object position feature at time $t$, respectively. Input vectors  \textcolor{black}{$ u^p_\mathbf{t} \in \mathbb{R}^{T\times 3}$ and $u^e_\mathbf{t} \in \mathbb{R}^{T \times 12}$ } indicate a time series of the observed robot pose tracking data and eye-tracking data vectors. $T$ is the length of the time window, and the vector $\mathbf{t}$ represents the previous $T$ steps of the temporal segment from $t$. }

 \textcolor{black}{The input-output mappings of each single modality are defined as: $v^*_t = g_*(u^*_t) \in \mathbb{R}^{d_f^*}$, where $g_{*}, \{* | i, p, o ,t\}$ represents the 2D or 3D shaped single-modal CNN structure which receives $u^*_t$ as the input, and the $v^{*}_t$ is the output of this structure.  $v^{*}_t$ also represents the feature vector extracted for the single modality. }

We concatenate the outputs from the various CNNs to create a combined feature vector, using $\boldsymbol{v_t} = [v^i_t:v^p_t:v^o_{t}:v^e_t ]$.
A fully connected layer combines the feature vector to produce a higher-level representation: $\boldsymbol{h_t} = \sigma(\mathbf{W}_f \boldsymbol{v_t}+ \mathbf{b}_f)$, where $\sigma(*)$ is the function of full connected layer; $\mathbf{W}_f$ is the weight parameter vector; and $\mathbf{b}_f$ is the bias parameter vector. 
We then feed the output from the fully connected layer into the last hidden state to compute the final output that represents the intended grasping position, by considering $\boldsymbol{o_t} = f(\mathbf{W}_o \boldsymbol{h_t}+ \mathbf{b}_o)$.

\section{System implementation}
This section describes the system overview, the proposed MMIPN method, user interfaces, and user interactions. These key components enable human teleoperators to perform grasping tasks in immersive environments. To generate decent immersive experiences and seamless operations, a laptop equipped with an NVIDIA RTX 4060 graphics card, 32 GB of RAM, and an i9-13900H runs the VR teleoperation user interface.

\subsection{System Overview}\label{system_overview}
\begin{figure}[hpt!]
\centering
\includegraphics[width=.8\columnwidth]{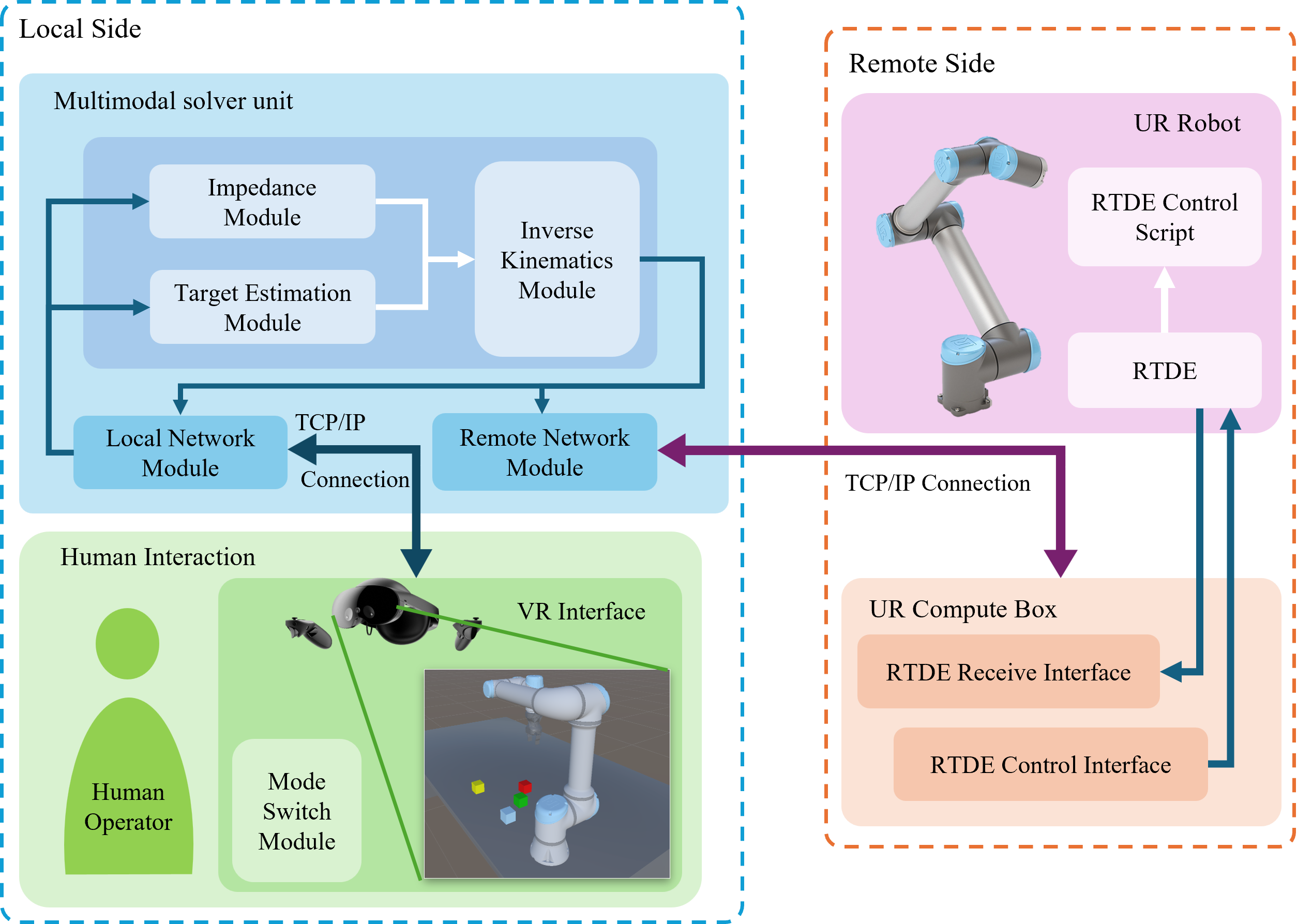}
\caption{The overview of the implementation of the \textcolor{black}{\textit{MIRAGE}} framework.}
\label{fig:fig_3}
\end{figure}

\cref{fig:fig_3} depicts the system overview. 
On the remote side, the robot model, namely UR5e, serves as the remotely operated robot with a 6-DOF robot arm. The robot has a local host computer, UR Compute Box \cite{onrobot2018}, to manage the remote and local messages under the proprietary RTDE communication protocol. An adaptive gripper, Robotiq 2F-85, is installed at the end of the UR5e.  The lower computer on the UR robot receives the control signal under the RTDE protocol, automatically generates the RTDE script, and sends the robot's underlying drive instructions. A TCP/IP connection is established with cables to bridge the communication between the remote and local sides to realize the teleoperation framework. 

The local side hosts both the human interaction and the multimodal solver unit of \textit{MIRAGE}. The VR headset, Meta Quest Pro HMD, was applied to present the teleoperation VR interface, which can acquire real-time eye-tracking data during the operation process. This system also incorporates Meta Quest Touch Pro Controllers to capture human operators' motion, allowing user interaction with immersive interfaces.

The multimodal solver unit is responsible for most of the calculation work. The virtual environment scripts integrate the task switch module. Furthermore, the Impedance module designed in Section~\ref{admittancecontrol}, the Inverse Kinematics module introduced in Section~\ref{IK}, and the Target Estimation module integrated with the MMIPN method proposed in Section~\ref{MMIPN} run in separate Python scripts. The information between the VR interface server and the Inverse Kinematics module or the Impedance module is a local TCP/IP connection. In contrast, the multimodal data transfer between the Target Estimation module and the VR interface is achieved through write and read local file storage, as well as through a local TCP/IP connection to transmit processing commands.

When the operator starts to manipulate the VR interface, the virtual impedance is activated, and the pose of the end effector is calculated based on the APF of the targets. The grasp position is calculated by the MMIPN module if the grasp command is given through the corresponding button pressed by the operator. These two functions could be switched to be enabled or not independently for the experiment settings.

\subsection{MMIPN Dataset and Model Training}

In order to test the suggested MMIPN before adding it to our system, we first created a multimodal grasp estimation dataset in the system shown in Section ~\ref{system_overview}. A total of three subjects were invited, each contributing 125 sets of dataset samples.
The data of each set includes the current field of view image ($\rm{H}\times \rm{W} \times \rm{C}=160\times160\times3$) when the subject moves the robot arm to align the operation and decides to perform the grasping action, the binocular 6-DOF eye tracking data of the first three frames, the 3-DOF robot end position of the first three frames, and the available object position group in the current observation environment.
The time window for the time series data is set to $T=3$.
The position of the actual target object currently specified by the task is collected as the ground truth in the dataset.

Regarding training settings, the network utilizes a Mean Absolute Error (MAE) regression loss function to effectively quantify the difference between predicted grasping positions and actual ground-truth positions. This measurement is crucial for guiding the optimization process. The Adam optimizer is employed in the training process to facilitate convergence and enhance the model's performance over time. To ensure robust performance and prevent overfitting, the complete dataset is divided into training and testing subsets in a usual ratio. Such an allocation enables effective monitoring of real-time test data during training, ensuring that the model's generalization capabilities are rigorously evaluated. A batch size of 8 is selected, and the training is conducted over a total of 100 epochs, providing ample opportunities for learning while allowing for sufficient validation at each stage. The learning rate selected during the model training phase is 0.005.

\begin{table}[h]
    \centering
    \footnotesize
    \caption{Ablation Study Results.}
    \begin{tabular*}{\columnwidth}{@{\extracolsep{\fill}} l l c c c @{}}
        \toprule
        Model & Modality & \makecell[c]{MAE\\(mm)}  & \makecell[c]{MSE\\($\rm{mm^2}$)} & MAPE\\ \midrule
        MMIPN & all modalities & \textbf{15.2} &  \textbf{409.6} &\textbf{11.79\%}  \\
        MMIPN-1 & without gaze & 442.8 & 319663.1 & 880.71\% \\
        MMIPN-2 & without robot's action & 30.6 &1437.1 & 36.19\%  \\
        MMIPN-3 & without object positions & 21.2 &702.3  & 19.95\%  \\
        MMIPN-4 & without observing image  & 20.8 &687.6  & 18.11\% \\ \bottomrule
    \end{tabular*}
    \label{tab:ablation_study}
\end{table}

In order to test the robustness of grasp position estimation using each modality channel of our proposed MMIPN network method, we designed a modality ablation study during the training phase. 
By systematically removing a single channel input from the complete MMIPN model, we obtained four pruned models with 3-channel inputs. We trained them using the same network training parameters, configurations, and the same training and testing subsets. The \textcolor{black}{MAE} result of the ablation study is shown in~\cref{tab:ablation_study}, \textcolor{black}{the Mean Squared Error (MSE) and Mean Absolute Percentage Error (MAPE) are also included in the result}. The complete MMIPN model, utilizing all four modalities, achieves the highest accuracy (MAE=15.2mm), indicating effective integration of multimodal inputs. Removing the eye-tracking modality (MMIPN-1) leads to the largest error in the estimation result, reflecting its essential role in providing the necessary human interaction information for intention recognition. In the absence of other modalities (MMIPN-2,-3, and -4), the estimation results demonstrated certain robustness, while the performance was still worse than that of the complete model (MMIPN).

The result of the ablation study illustrates that the chosen 4 modalities as MMIPN inputs show their significance at different levels to estimate the human-intended grasp target position. The complete MMIPN model shows the best performance; the trained complete model of MMIPN is utilized in our system. 

\subsection{System Interfaces}

\begin{figure}[hpt!]
\footnotesize
\centering
\includegraphics[width=.8\columnwidth]{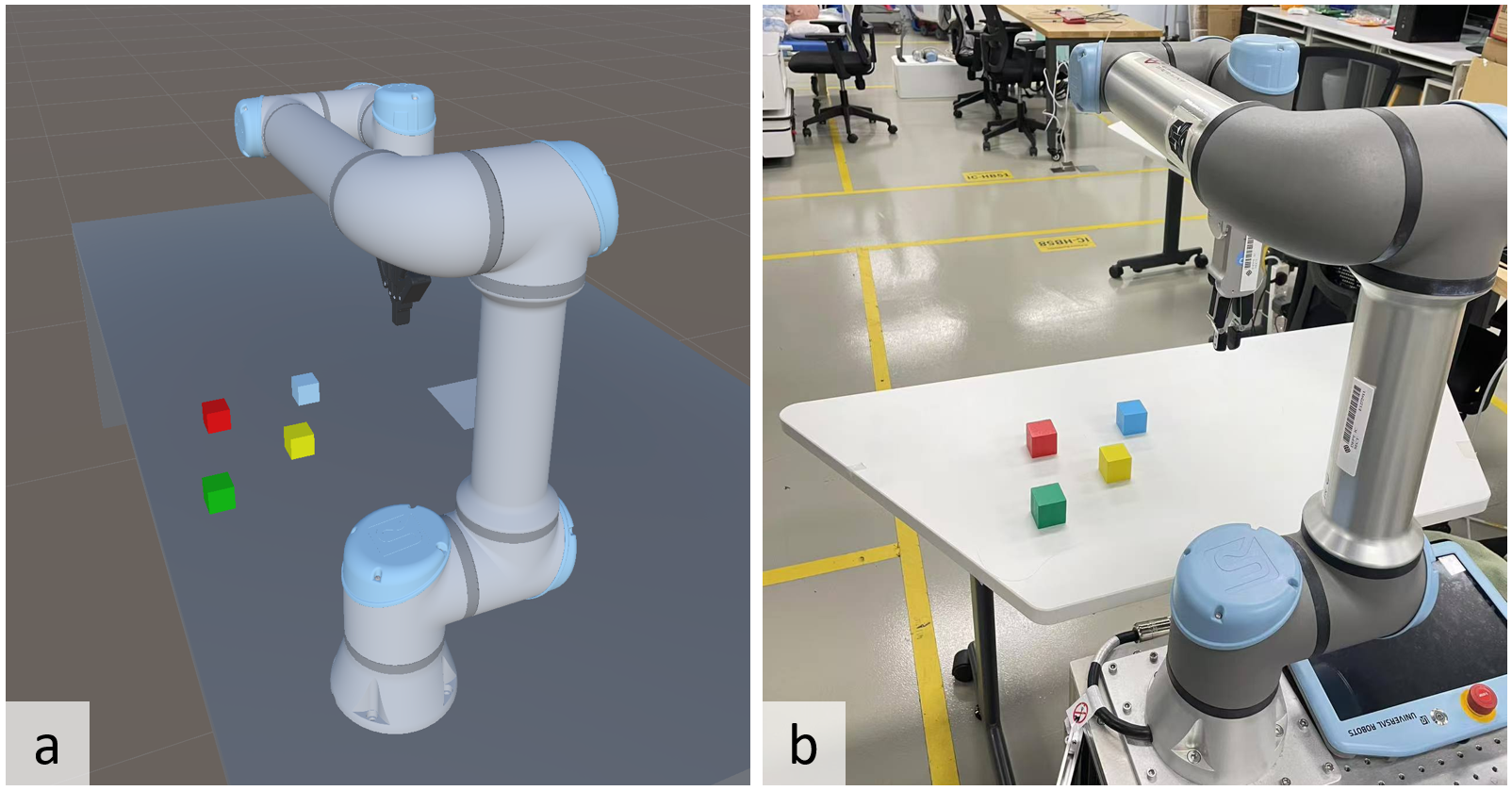}
\caption{The virtual-physical configurations: the comparable settings of robotic arms with grippers in the (a) virtual system vs (b) physical system. 
}
\label{fig:interface}
\end{figure}
The virtual environment is built in Unity (version 2022.3.34f1). The virtual scene includes a simple grey table, and the UR5e robot is installed on the table, similar to the laboratory setting, which is shown in ~\cref{fig:interface}. The Robotiq 2F-85 Gripper is fixed as the end effector of the virtual robot arm.  We use manufacturer-provided 3D models of the robot and end effector to ensure consistency between the virtual and the real environments.

The robot base is placed close to the user, and a basic folding configuration initially makes the robot face forward. This setting allows the operator to quickly understand the reflection between the human action input and the motivated robot motion. A prompt bar is designed and placed in the upper left corner of the user's field of view (in a VR environment) to provide some task-related information if necessary. The controllers used are shown inside the VR environment, allowing users to verify their interaction buttons across various control modes, which will be elaborated upon in the subsequent section.
The objects for teleoperation tasks will also be placed on top of the table.


\subsection{User Interaction}

The standard controllers of Meta Quest Touch Pro are used as mentioned in Section~\ref{system_overview}, and the right controller is applied as the human input device,
while the left controller also needs to be held in hand to ensure the stable tracking of the controllers by the Meta Quest Pro HMD. During the interaction in the system, operators will experience two different modes of control in the shared control teleoperation framework. And the proposed implicit assistance methods, \textbf{VA} in Section~\ref{admittancecontrol} and \textbf{intention perception} in Section~\ref{MMIPN}, are implemented in separate modes.

The first mode is \textbf{direct shared control}, where the actions of the human operator directly motivate the robot's motion, and the proposed model of \textbf{VA}  is integrated into the direct shared control scheme. 
When the operator continuously presses the `Grip Button', the position tracking will be kept recorded and solve the desired actual position of the robot end following the calculation process in Sections~\ref{shaed_control} and~\ref{admittancecontrol}, then the robot motion is driven by the joint angles command which is obtained through the calculation instruction in Section~\ref{IK}. 
The operator can control the movement of the robotic arm's end to maintain a basic alignment with the direction of the arm's movement. 
This mode uses the incremental direct teleoperation control, and the release of the `Grip Button' on the controller represents the `off' of the switch function in  \cref{equ:telemapping}, which helps the operators to extend the workspace when they reach the limit of a certain direction.
The system's loop latency between generating human motion inputs and driving the robot actuators is measured as $56.0 \pm 3.8$ ms, including network transmission, inverse kinematics calculation, and VR rendering or actuator execution in the loop under this mode.

In the \textbf{supervised shared control} mode, the robot motion is guided by the operator's instruction, while its motions are planned automatically with the support of \textbf{MMIPN}. 
Once the operators press the `A Button' on the right controller to give the grasping command to the robot, the eye-tracking data,  the first-person view observation image, the time sequence robot motion data and the positions of the objects can be obtained in the scene are all collected and delivered to the target estimation module to estimate the desired grasping position. On our hardware, the MMIPN computes $249.3 \pm 7.1$ ms, which only occurs in the loop once in one grasp decision. Then, the robot executes a self-planned path according to this estimation to complete the grasp action. 
\textit{MIRAGE} primarily operates in direct shared control mode. Once switched to supervision shared control mode, it remains in that mode until the queue of self-planned actions has been fully executed.

\section{Study Design} 
\subsection{Conditions}\label{conditions}
The two methods we proposed in \textcolor{black}{\textit{MIRAGE}} offer implicit operational assistance at distinct stages of HRI during teleoperated grasping. Consequently, our primary focus is to assess the extent to which these methods enhance operational performance during the grasping process. Additionally, we aim to evaluate the specific characteristics of the operational experience that are improved by each method. To this end, we identify two key factors to evaluate, along with the primary similarities and differences they exhibit within the system, which are also illustrated in ~\cref{fig:teaser}: 
\\
\textcolor{black}{\textbf{Factor 1: Virtual Admittance (VA)}. \textcolor{black}{When VA is enabled, the system generates implicit guidance through artificial potential fields surrounding objects, creating a sensation akin to gravitational pull as the robot approaches an object. The robot's trajectory, under shared control with the operators and influenced by virtual force, is smoothly adjusted, allowing the end of the robot arm to be positioned closer to the object than the distance dictated by the operator's input.} Without the VA, the distance of the operator's movement is directly reflected on the robot. 
\\
\textbf{Factor 2:  Intention Recognition (MMIPN)}.  \textcolor{black}{Intention recognition} driven by MMIPN affects the experience in the supervised shared control mode to grasp the object. With MMIPN, the robot will generate its path according to the estimation from MMIPN, and the operator will observe that the robot's grasping position is perhaps not on the vertical projection of the end of the robot arm. Without MMIPN's assistance, the robot will plan its path to descend vertically to grasp objects.} 
Accordingly, we use a $2\times2$ factorial design~\cite{HAERLINGADAMSON202090twobytwo} to examine the experimental systems and conduct experiments in the following four conditions: \textcolor{black}{ Condition 1: The system without assistance (None, Baseline);  Condition 2: The system with the VA assistance (VA Only);  Condition 3: The system with the assistance of  \textcolor{black}{intention recognition} (MMIPN Only);
and Condition 4: The system with both VA and \textcolor{black}{intention recognition} assistance (\textcolor{black}{\textit{MIRAGE}}: Combination of VA and MMIPN). }

These four experimental conditions, as represented by system capabilities and features, share the identical virtual scene and task arrangement, with the only difference in the condition setting of the two factors: VA and MMIPN. \textcolor{black}{The value of mapping parameters in ~\cref{equ:telemapping} and in ~\cref{equ:apfmodel1} are set to $k_m = 0.3$ and $k_i = 0.1$. The MMIPN model is utilized with consistent parameters across different conditions and participants, without individual training.} To evaluate the operators' performance and experience in different conditions, the user study is a within-subject design~\cite{charness2012withinsubjects}, and the counterbalanced design by the Latin Square technique is used for alleviating carryover effects in within-subject designs.

\subsection{Task Design}
Participants were asked to complete the same tele-grasping task under four different conditions. 
In each condition, the scene was presented from a \textit{Perspective View}. The \textit{Perspective View} enables users to maintain a panoramic perspective and gain a comprehensive understanding of the task scenarios. However, the \textit{Perspective View} inherently presents depth perception challenges, making it difficult for users to estimate the depth between the robotic arm and the target object in the virtual environment. 
Each participant stood next to a virtual table positioned to their right-hand side. The task comprised 10 blocks, with each block containing four trials, resulting in a total of 40 trials per participant.

\begin{figure}[ht!]
\centering
\includegraphics[width=.8\columnwidth]{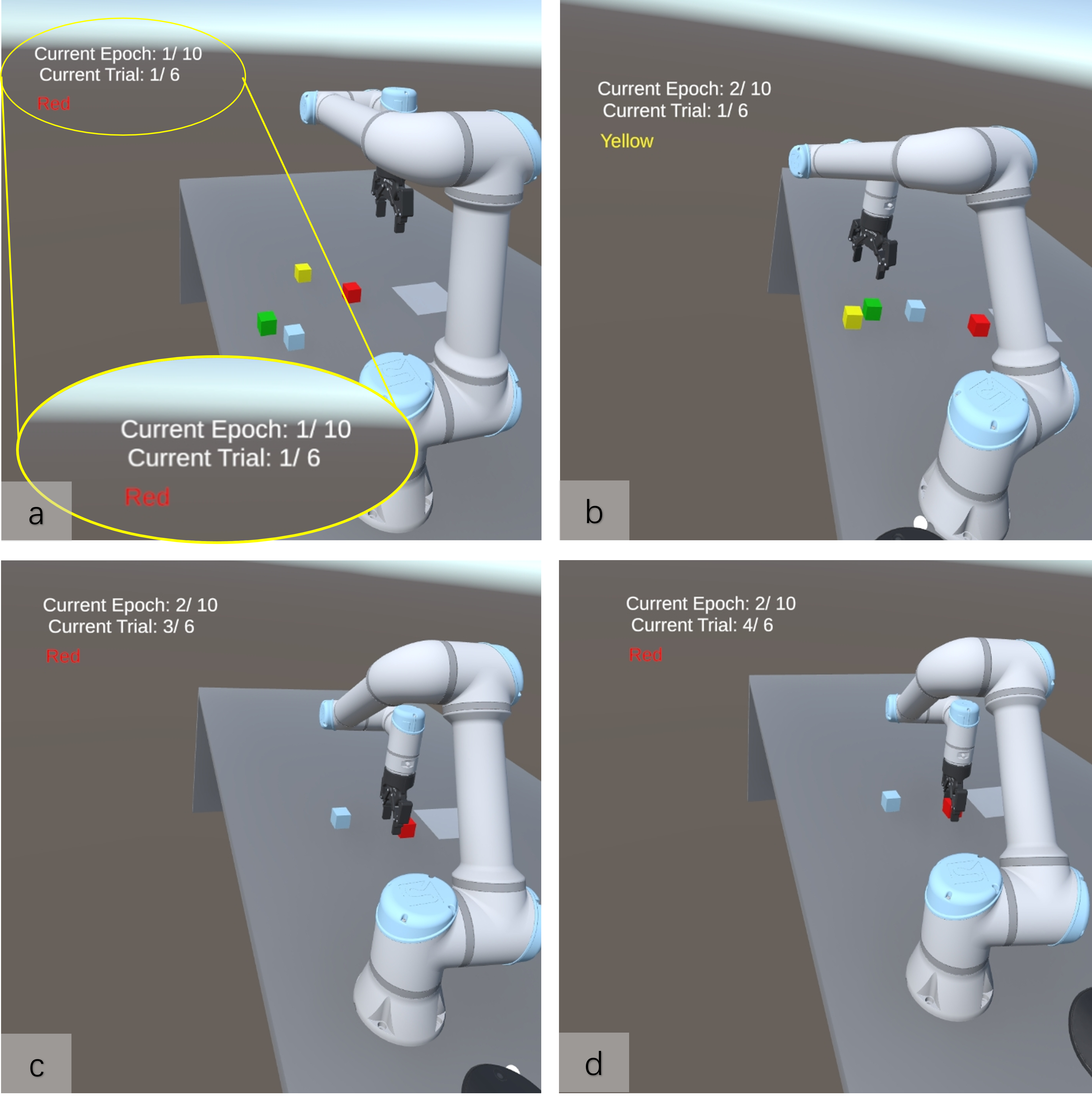}
\caption{The task-related information is given in the prompt bar on the user's interface (a), including the current target (described in colors), the current attempt (Trial), and the current block (Epoch).  In the task, the objects' positions are randomly refreshed in a new block, comparing (a) and (b). After a failed attempt in one block, if the current number of attempts in this block has not reached the 6th attempt (max), the target remains until the grasp is successful (d). }
\label{fig:taskdemo}
\end{figure}

At the beginning of each block, four colored cubes were randomly placed on the virtual table, serving as targets for the upcoming trials. \textcolor{black}{The grasping order was also generated randomly in each block. On each trial, the grasping targets were indicated in this order, and participants had to grasp the corresponding color block according to the prompt (~\cref{fig:taskdemo} [a]).} Within each block, participants were allowed a maximum of six grasp attempts. If they exceeded this limit, the task would automatically move on to the next block. The interface of the task and the key features of the task arrangement are shown in~\cref{fig:taskdemo}.

\subsection{Procedure}

At the beginning of the experiment, participants were asked to fill in the self-reported demographic information. 
Then, participants were given a 5-minute tutorial to familiarize themselves with the VR interface, interaction methods for controlling the virtual robot and manipulating objects, and the task-related information presented via the prompt bar. 
The formal experiments were conducted to ensure that the participants fully understood the system's operation and task instructions after the tutorial. Each participant experienced all four systems across 10 blocks. After a complete experience in one system, the participants were asked to fill out 3 post-test questionnaires and take a 2-minute rest while sitting in the chair. After completing all system tests, each participant took part in a 5-minute semi-structured interview to further collect their subjective feelings and feedback on the different conditions of the system.

\subsection{Participants}
We recruited 16 participants (11 female and 5 male) from local universities, aged between 23 and 27 years (M = 24.47, SD = 1.38). The average height of the participants was 169.33 cm (SD = 6.66 cm). 3 participants reported they have no experience with VR, and familiarity of the participants with VR technology was measured on a 5-point scale, with 3 indicating no familiarity at all; the mean VR familiarity score was 2.67 (SD = .73). 9 participants reported they have previous experience with robots, and familiarity of the participants with robots was also measured on a 5-point scale, with 1 indicating no familiarity at all; the mean robot familiarity score was 2.35 (SD = .57). All of the participants are right-handed. The study was approved by the university's ethics review board (Application Number: HSEARS20250328007).

\subsection{Measurements}

We employed both quantitative and qualitative analyses to evaluate the user's task performance and experience across these two factors. The specific metrics are listed in the following sections.

\paragraph{Objective Metrics}
To quantitatively evaluate the task performance, we record the outcome (success or failure) of each grasp attempt. Performance metrics were calculated across 10 experimental blocks, including grasp success rate ((Number of successful grasps / Number of grasp attempts) $\times$ 100\%) and absolute successful grasp count. We defined two specific performance categories: \textbf{Bad Blocks} (failure to grasp 4 objects within 6 attempts in a single block) and \textbf{Perfect Blocks} (completion using exactly 4 attempts in a single block).
Additionally, we measured temporal efficiency by recording operation duration at both the individual attempt and complete block levels, allowing calculation of mean grasp time and mean block completion time. 
To assess operator motion behavior, we tracked participants' hand kinematics by capturing controller position data during direct shared control operations at a $10 Hz$ sampling rate. From these trajectory data, we derived several quantitative metrics: total movement distance, movement efficiency (distance per successful grasp), and movement velocity (distance per unit time (s)). 

\paragraph{Subjective Metrics}
For subjective assessments,  we administered the NASA-TLX Scale (NASA-TLX) to measure subjective perceived workload~\cite{Hart2006Nasa}, the User Experience Questionnaire (UEQ) to evaluate user experience with the interactive product~\cite{ueq2017}, and the Slater-Usoh-Steed (SUS)  to measure user presence~\cite{SUS1998}. We conducted the post-study interview to understand their preferences and their reasons.

\section{Results}

\begin{figure*}[h]
\centering
\includegraphics[width=.9\linewidth]{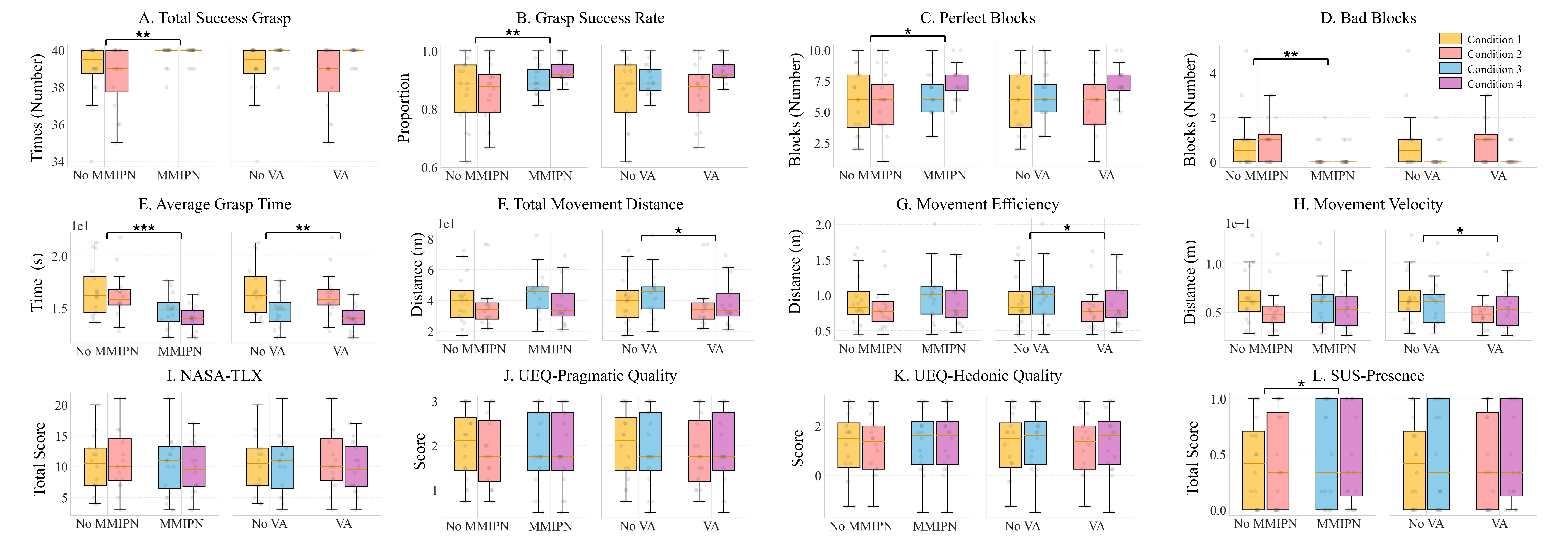}
\caption{Box plots of objective metrics (A-H) and subjective questionnaires (I-L) across each condition. \textcolor{black}{($p < .05$(*), $p < .01$(**), $p < .001$(***))} 
}
\label{fig:result1}
\end{figure*}

The collected data are presented in ~\cref{fig:result1}. To evaluate the results within a two-way within-subjects design, we performed a two-way repeated measures ANOVA (RM-ANOVA) for statistical analysis~\cite{edwards1950experimental_ANOVA}. Prior to running RM-ANOVA, we assessed normality using the Shapiro-Wilk test and tested the sphericity assumption with Mauchly’s test. When sphericity was violated, the Greenhouse-Geisser correction was applied to adjust the degrees of freedom. If the normality assumption was not satisfied, we adopted the Aligned Rank Transform (ART) ANOVA~\cite{wobbrock2011ARTANOVA} as a nonparametric alternative. \textcolor{black}{For the significance of main effect, no post hoc analyses were required because both factor has only two levels in our study as introduced in \cref{conditions}. We conducted the post hoc analyses to further investigate the simple effects when a significant interaction effect was observed.} The post hoc pairwise comparisons were performed using paired t-tests when parametric assumptions were met, and the Wilcoxon signed-rank test was used otherwise, with a Bonferroni adjustment applied to control for multiple comparisons.  These procedures were consistently applied across all analyses to ensure the robustness and reliability of the results.

\subsubsection*{Task Completion and Grasp Success Rate}

The ANOVA results showed that MMIPN's assistance significantly enhanced grasp performance across multiple metrics. Specifically, it significantly increased the successful grasp count ($F(1,15)=12.31$, $p<.01$, $\eta^2=.151$), while neither VA ($p=.680$) nor the interaction between factors ($p=.584$) reached statistical significance. Similarly, there was a significantly higher grasp success rate with the assistance of MMIPN ($F(1,15)=9.63$, $p<.01$, $\eta^2=.151$). 
Further analysis demonstrated that MMIPN significantly increased the occurrence of \textbf{Perfect Blocks} ($F(1,15)=19.14$, $p<.05$, $\eta^2=.064$) while VA did not show a significant effect ($p=.349$) and the interaction between the two factors approached but did not reach significance ($p=.066$). In contrast, the absence of MMIPN significantly increased the frequency of \textbf{Bad Blocks} ($F(1,15)=8.27$, $p<.01$, $\eta^2=.134$), with neither VA ($p=.901$) nor the interaction ($p=.901$) showing significant effects.

\subsubsection*{Temporal Efficiency}

ANOVA results revealed that both VA ($F(1,15) = 8.91$, $p < .01$, $\eta^2 = .026$) and MMIPN ($F(1,15) = 25.50$, $p < .001$, $\eta^2 = .221$) assistance significantly reduced the mean grasp time. However, the interaction between the two types of assistance did not \textcolor{black}{significantly affect} temporal efficiency ($p = .62$). Additionally, no significant differences were observed in mean block completion time under either assistance condition (VA: $p = .51$; MMIPN: $p = .20$). 

\subsubsection*{Movement Performance}

With the assistance of VA, the total movement distance during the task was significantly reduced ($F(1,15)=4.66$, $p<.05$, $\eta^2=.024$). Movement efficiency improved by VA ($F(1,15)=5.02$, $p<.05$, $\eta^2=.022$), and movement velocity also showed a significant increase under VA support ($F(1,15)=7.71$, $p<.05$, $\eta^2=.038$). In contrast, MMIPN's assistance did not lead to significant improvements in movement-related metrics (total movement distance: $p=.30$; movement efficiency: $p=.06$; movement velocity: $p=.47$). No significant interaction effects were observed between the two types of assistance across total movement distance ($p=.46$), movement efficiency ($p=.92$), or movement velocity ($p=.359$). 
VA significantly optimized movement trajectories across all metrics ($\eta^2=.022-.038$), confirming its effectiveness in improving movement efficiency through impedance adaptation. MMIPN’s non-significant effects reinforce its role as a perceptual aid rather than a motion optimizer.
However, the effect of MMIPN on movement efficiency approached significance ($p = 0.06$), which may be attributed to its capability to estimate grasp alignment, thereby reducing operator hesitation during positioning.

\subsubsection*{Subjective Perceived Workload (NASA-TLX)}
The results showed that the \textcolor{black}{VA} assistance has no significant impacts on the Mental demand ($F(1,15)=.41$, $p=.53$), Physical demand ($F(1,15)=.006$, $p=.94$), Temporal demand ($F(1,15)=.90$, $p=.36$), Performance ($F(1,15)=.91$,  $p=.36$), Effort ($F(1,15)=.19$, $p=.66$) and Frustration ($F(1,15)=.08$, $p=.77$). We also did not find a significant effect of the assistive MMIPN on the Mental demand ($F(1,15)=.39$, $p=.54$), Physical demand ($F(1,15)=1.25$, $p=.28$), Temporal demand ($F(1,15)=.003$, $p=.96$), Performance ($F(1,15)=.029$, $p=.867$), Effort ($F(1,15)=3.01$, $p=.103$) and Frustration ($F(1,15)=.30$, $p=.58$).

\subsubsection*{User Experience Questionnaire (UEQ)}
We pre-processed the data for the UEQ questionnaire according to the instructions \cite{laugwitz2008UEQconstruction} for three main conditions: the Pragmatic Quality, the Hedonic Quality and the Overall Satisfaction. Both with the assistance of VA ($F(1,15)=.36$, $p=.559$) and MMIPN ($F(1,15)=.30$, $p=.589$), the system shows no significant difference in the Pragmatic Quality. 
The VA in the system has no significant impact on the Hedonic Quality ($F(1,15)=.06$, $p=.814$) and Overall Satisfaction ($F(1,15)=.48$, $p=.497$). With MMIPN, the system has shown a better mean value on the Hedonic Quality and Overall Satisfaction, but not found a significant difference (Hedonic Quality: ($F(1,15)=2.97$, $p=.105$), Overall Satisfaction: ($F(1,15)=2.56$, $p=.130$)).

\subsubsection*{Slater-Usoh-Steed (SUS)}
We first followed the instruction~\cite{usoh2000SUSusing} to score each questionnaire, and the results show that the assistance of MMIPN has a significant impact on presence ($F(1,15)=5.94$, $p <.05$, $\eta^2=.86$). The VA has no significant impacts on presence ($F(1,15)=.38$, $p=.54$). Also, there was no significant impact among the interaction of two factors ($F(1,15)=.59$, $p=.45$). The results imply that MMIPN’s significant presence enhancement ($\eta^2=.86$) underscores the importance of such support in VR-based teleoperation. By providing a grasp estimation, MMIPN likely reduces cognitive dissonance resulting from the alignment error under a single view.
\subsubsection*{Qualitative Feedback} \label{interviewfeedback}
Most participants (N=9) reported having the best experience with the system in Condition 4 (\textit{MIRAGE}). They expressed their fondness for the system in Condition 4 in interviews, as one said, \textit{``I feel more confident in both manipulating and grasping when I feel both assistance working together to assist me.''}

Most participants (N=9) preferred the system with VA involvement (Conditions 2 and 4). A few participants (N=3) reported their maladjustment under these conditions, as one participant said in the interview: \textit{``I lost my sense of control, so I operated (the gripper) from a distance from objects to avoid being affected.''} Another participant mentioned that \textit{``The system needs to adapt to the feel.''}
Although most participants (N=10) liked the system involved MMIPN (Conditions 3 and 4)  and reported that MMIPN improved their grasping success rate in completing the task, two participants said that under these conditions, the position estimates provided by the system were sometimes inconsistent with their intentions.
Additionally, the participants discussed their different feelings about the two factors, with one participant stating, \textit{``The gravitational effect makes the manipulation more flexible, but it has little effect on the success rate, and the prediction estimate is very helpful in completing the manipulation task.''}

Furthermore, four participants mentioned that their standing position and single viewpoint affect their performance in alignment: \textit{``In a single perspective, object alignment will have parallax, which is confusing.''}, which mainly explains the failure and worst performance in the baseline (Condition 1), \textit{``I felt most difficulty in this system (Condition 1). For objects placed at the edge or in a special position, aligning them with the gripper is difficult. And there is parallax in a single view.''} Another participant described her feeling in Baseline as, \textit{``I feel helpless without any help.''}

\section{Discussion}
\paragraph{Virtual Admittance: Streamlining Motion through Implicit Guidance} VA demonstrates its benefits in optimizing gross motion trajectories by reducing the operator's path length compared to unassisted teleoperation. By generating virtual guiding forces based on an artificial potential field, VA implicitly senses the user's preferred target and guides the user towards the target object without restrictive force feedback, thus preserving the actor's agency. This is in line with a prior study by Sun et al.,~\cite{sun2024development}, which emphasized the effectiveness of impedance-based guidance in harmoniously coordinating human movement intent and assistance with robot autonomy in a shared control framework. 
Meanwhile, VA shows no significant enhancement in Grasp Success Rate ($p>.05$), which is mainly because it provides coarse spatial guidance donated by the attractive force from several nearby objects, rather than precise gripper alignment with a single target.
In addition, unlike haptic shared-control systems~\cite{abi2019haptic, adjigble2019IROSassisted}, which physically constrain operators via force feedback, VA’s visual guidance is non-contact and may avoid inducing resistance or discomfort. 
However, VA is not robust to subjective workloads ($p>.05$) and shows an absence of effect, suggesting that it primarily addresses spatial efficiency rather than perceptual challenges. 
For future work, the study can be extended to more complicated tasks and lengthy operations, which could potentially uncover different user behaviors and workload patterns.

\paragraph{\textcolor{black}{Intention Recognition}: Gaze as the Key of Disambiguation}

MMIPN's superior accuracy (MAE=15.2mm), compared to non-gaze models (MAE=442.8mm), a 96.6\% improvement underscores the gaze's pivotal role in disambiguating intent within 3D scenes, aligning with the result of Li et al., ~\cite{li20173}, who demonstrated the potential of 3D gaze in object localization to support perceptual functions essential for operation planning, akin to human visuomotor behavior. Notably, as shown in the ablation results in~\cref{tab:ablation_study}, robot motion data (MMIPN-2) emerged as the second most critical modality, highlighting its strong correlation with human actions. This finding aligns with Wagner et al.~\cite{wagner2024UISTeyehand}, who demonstrated that incorporating gesture data can enhance object positioning accuracy in VR-based HRI by mitigating errors caused by gaze input jitter and slippage.
Unlike previous gaze-based systems that depend on explicit dwell-time selection~\cite{li20173,pan2024gaze}, MMIPN infers user intent implicitly by leveraging natural gaze behaviors.

\paragraph{User limitations: Parallax and Perspective}
Visual feedback is essential in teleoperation, and demonstrations using VR’s stereoscopic vision have greatly enhanced the sense of presence. It is important to note that, in VR-based HRI, operators often struggled with perspective-induced parallax in single-view setups, as indicated by post-test interviews in Section \ref{interviewfeedback}. 
Our result aligns with the findings of Barrera et al. ~\cite{Barrera2019CHIstereodesplay}, who discovered that the users' performance was worse when moving along the depth axis compared to other axes in a virtual display, though this issue was not observed in a real-world setting during their experiment.  This problem is particularly evident when performing alignment operations, and the target estimation of MMIPN performs well in our system to overcome this difficulty. As mentioned in the previous section, the binocular gaze used in MMIPN also implies depth information, which complements the visual observation and motion information modalities, thus making up for the errors of human eye observation. 
Additionally, considering that Barrera et al. found that there is no obvious depth error in real-world manipulation, it is reasonable that our method enhances the sense of presence by weakening the influence of parallax.

We used a third-person view to arrange the setting in the VR interface, which allowed the robot arm and target objects to be fully visible in the scene for user reference. 
Filthaut et al. found that, compared with the first-person perspective manipulation, the third-person perspective operation in HRI teleoperation gives the operator a stronger sense of presence because it can provide complete environmental information and also leads to better performance in HRI under the telepresence context~\cite{filthaut2024iSpaRothirdpersonview}. However, during the post-study interview, one participant expressed discomfort with the fixed third-person viewpoint coupled with the robot, indicating a preference for operating from the perspective decoupling with the robot~\cite{walker2023virtual}. This observation does not contradict the benefits of third-person perspectives; rather, it highlights that individual cognitive habits and spatial preferences may influence viewpoint choices in unconstrained teleoperation tasks. This warrants further investigation in future work.

\paragraph{Assistive Modalities: Transparency and Sense of Agency}
The four experimental conditions we established across two factors (Section ~\ref{conditions}) show some interesting insights. Condition 1, which provided no assistance, served as the baseline and resulted in the lowest overall performance. However, some participants performed well within this unassisted system and expressed satisfaction with it during the interviews. In Condition 2, which included only virtual guidance, several participants expressed discomfort and a perceived loss of control, as highlighted in the feedback presented in Section ~\ref{interviewfeedback}. Additionally, Conditions 3 and 4 received two pieces of feedback regarding estimation errors from MMIPN, which caused confusion.

These issues may stem from the implicitness of assistance methods, which leads to the users' lack of understanding about the technology behind these assistance methods. This reduces the transparency of the HRI process on technical clarity. In contrast, human intention recognition based on multimodality aligns with human intuition in terms of interaction mode and performance, making it easier to be observed and predicted. This level of transparency helps ensure a more positive user experience and task performance in our studies. 
Moreover, transparency in HRI significantly impacts users' sense of agency (SoA)~\cite{GALLAGHER2012PIISOA}, which can further influence their confidence and overall performance in using the technology. The concept of transparency is multifaceted and warrants further exploration~\cite{alonso2018transparencysystem}. The implications of transparency on user experience and efficiency deserve additional design and research focus.

\paragraph{Limitations of Our Study}
Our system (Section ~\ref{system_overview}) is subject to a limited size of dataset, and the data reflects the task of grasping among multiple cubes without stacking. The MMIPN method is validated to be robust across this dataset and shows its effectiveness in the user study. These results give preliminary evidence to illustrate the MMIPN's ability to infer human intention in multi-object grasping. \textcolor{black}{While the current implementation demonstrates promising results in controlled settings, its performance in more complex scenarios involving multi-type objects, object-stacking configurations, or dynamic environments requires further validation. Future work will systematically evaluate the framework's adaptability in such scenarios, which are crucial for practical applications.}
\textcolor{black}{Furthermore, our participant cohort (N=16) primarily comprised university students. While suitable for initial testing, this limitation restricts the potential applicability of our findings to broader user groups with diverse skills, ages, or professional backgrounds. Future work will recruit more diverse participants to improve generalizability.}
\textcolor{black}{Additionally, MMIPN's high computational demands slightly affect VR rendering quality on our experimental devices. Several users reported mild discomfort while using the system under the conditions with MMIPN, which may have impacted their experience. However, there were no reports of significant computational latency compared to other non-MMIPN conditions.} The results of the ablation study (\cref{tab:ablation_study}) can reduce the computational demands by pruning the model's input channel while still needing more in-depth investigation.
Moreover, the \textcolor{black}{\textit{MIRAGE}} framework focuses on providing implicit assistance in the virtual environment based on natural human interfaces, like body motion and gaze. No visual cues or interactable interface are implemented explicitly in the system; the consideration is that the high-interactivity control systems (HICS), e.g., using motion mapping, generally result in better performance and lower cognitive load than low-interactivity control systems (LICS), e.g., using pointing interaction, in HRI tasks~\cite{nenna_enhanced_2023}. Nevertheless, the evaluation of these two levels of HRI systems in VR still requires additional studies, and further research may investigate their characteristics in VR-based HRI.
The role of gaze in shared control requires further exploration. Also, the adaptability and personalization of the VA guidance in \textcolor{black}{\textit{MIRAGE}} could be enhanced by introducing user-customizable parameters and integrating visible overlays into the framework.

\section{Conclusion}

This article serves as groundwork for a human-oriented VR-HRI method by integrating VA and multimodal intention recognition to address perceptual and motion challenges in multi-object teleoperation. The VA model reduced operator movement distance through artificial potential fields, while MMIPN implicitly completed target estimation by prioritizing gaze and increased the success rate in the task. The synergy of these components, validated by our user studies, demonstrates that implicit spatial guidance paired with gaze-driven intent disambiguation significantly enhances both efficiency and accuracy compared with the baseline, mirroring natural human visuomotor coordination.
In VR-based teleoperation, the system operates as a purposeful \textcolor{black}{\textit{MIRAGE}} that is designed to simplify the HRI process's complexity and user hindrance. 
By embracing the \textcolor{black}{\textit{MIRAGE}} design, where technology’s role is not to replicate reality but to enable the user's ability. Our HRI systems of virtuality with implicit assistance provide users with intuitive experiences in tele-operating robots. For future work, we will explore alternative feedback modalities and AI-assisted visualization, including extending the remote cues and improving the virtual assistance in the VR interface, to enable the \textcolor{black}{\textit{MIRAGE}} users with intuitiveness and performance.

\acknowledgments{
This research was supported by the Hong Kong Polytechnic University's Start-up Fund for New Recruits (No. P0046056), Departmental General Research Fund (DGRF) from HK PolyU ISE (No. P0056354), and PolyU RIAM -- Research Institute for Advanced Manufacturing 
(No. P0056767). Xian Wang was supported by a grant from the PolyU Research Committee under student account code RMHD.}

\bibliographystyle{abbrv}

\balance
\bibliography{VRHRI2024}
\end{document}